%% file: main.tex
\documentclass[10pt,twocolumn,letterpaper]{article}

\usepackage[pagenumbers]{cvpr} 

\input{preamble}
\definecolor{cvprblue}{rgb}{0.21,0.49,0.74}
\usepackage[pagebackref,breaklinks,colorlinks,allcolors=cvprblue]{hyperref}
\usepackage{multirow}
\usepackage{xcolor,colortbl}
\usepackage{algorithm}
\usepackage{algorithmic}

\def\paperID{****}
\def\confName{CVPR}
\def\confYear{2026}

\title{Trajectory-Diversity-Driven Robust Vision-and-Language Navigation}

\author{
 Jiangyang Li$^{1}$\quad
 Cong Wan$^{1}$\quad 
 Songlin Dong$^{1,2}$\thanks{Corresponding author.}\quad
 Chenhao Ding$^{1}$\quad
 Qiang Wang$^{1}$\quad \\
 Zhiheng Ma$^{2}$ \quad  
 Yihong Gong$^{1,2}$\quad
\\
$^1$Xi’an Jiaotong University \\
$^2$Faculty of Computility Microelectronics, Shenzhen University of Advanced Technology \quad
\\ 
}

\begin{document}
\maketitle
\input{sec/0_abstract}    
\input{sec/1_intro}
\input{sec/2_related_work}
\input{sec/3_method}
\input{sec/4_experiment}
\input{sec/5_conclusion}

{
    \small
    \bibliographystyle{ieeenat_fullname}
    \bibliography{main}
}

\input{sec/X_suppl}

\end{document}

%% file: sec/0_abstract.tex
\begin{abstract}
Vision-and-Language Navigation (VLN) requires agents to navigate photo-realistic environments following natural language instructions. Current methods predominantly rely on imitation learning, which suffers from limited generalization and poor robustness to execution perturbations.
We present \textbf{NavGRPO}, a reinforcement learning framework that learns goal-directed navigation policies through Group Relative Policy Optimization. By exploring diverse trajectories and optimizing via within-group performance comparisons, our method enables agents to distinguish effective strategies beyond expert paths without requiring additional value networks.
Built on ScaleVLN, \textbf{NavGRPO} achieves superior robustness on R2R and REVERIE benchmarks with \textbf{+3.0\%} and \textbf{+1.71\%} SPL improvements in unseen environments. Under extreme early-stage perturbations, we demonstrate \textbf{+14.89\% SPL gain} over the baseline, confirming that goal-directed RL training builds substantially more robust navigation policies. Code and models will be released.
\end{abstract}

%% file: sec/1_intro.tex
\section{Introduction}
\label{sec:intro}

\begin{figure*}[t]
 \centering
\includegraphics[width=\textwidth]{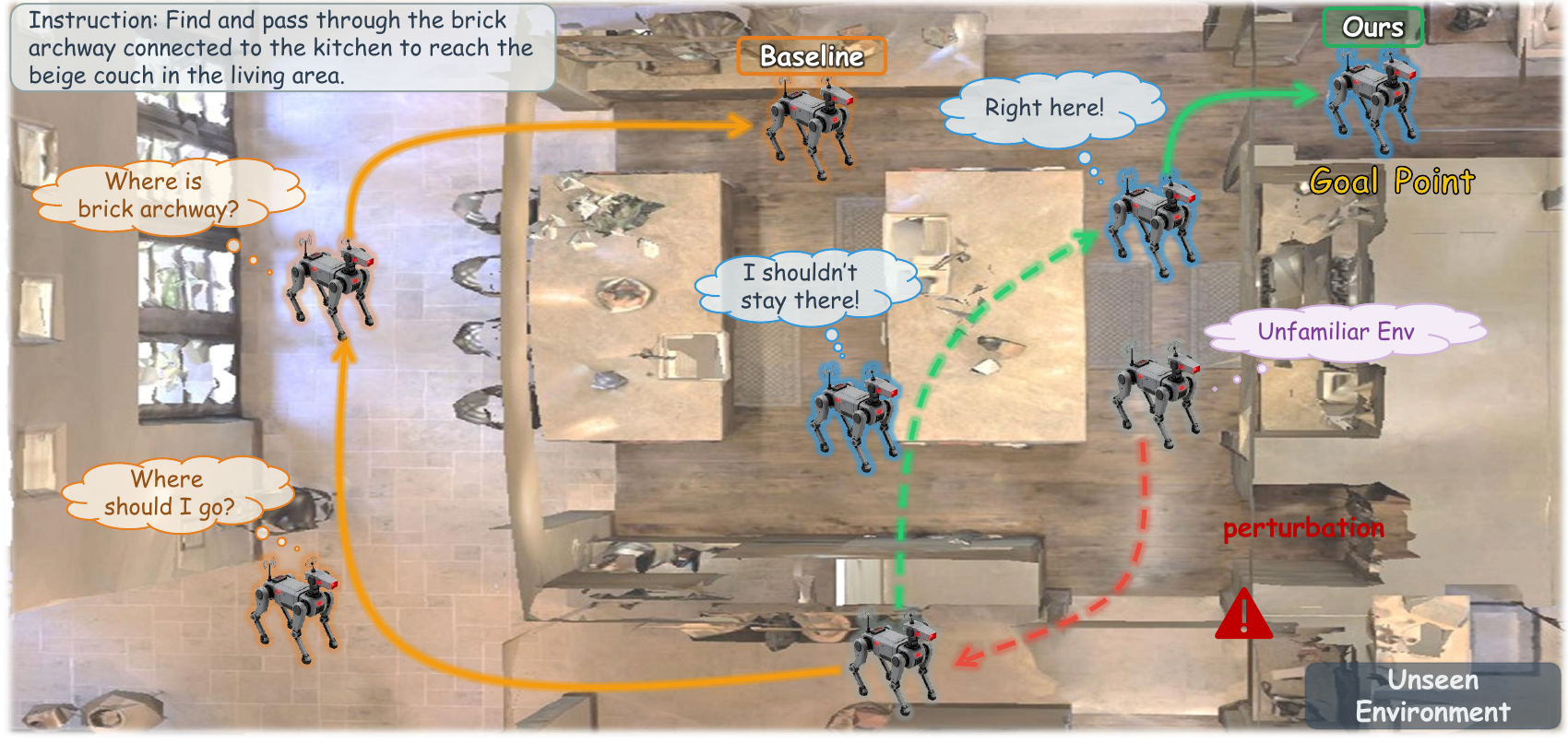}
\caption{Navigation behavior under early-stage perturbations in unfamiliar environments. The baseline IL agent struggles to recover from errors due to limited exposure to failed trajectories, often detouring or failing to reach the goal. Our NavGRPO agent learns from diverse rollouts, enabling robust error correction and successful navigation despite perturbations.}
\label{fig:teaser}
\end{figure*}

Vision-and-Language Navigation (VLN)~\cite{anderson2018r2r} is a core task in Embodied AI, requiring agents to autonomously navigate to target locations based on natural language instructions. This task is highly challenging as agents must ground abstract linguistic concepts to visual observations and perform multi-step reasoning to achieve navigation goals. Additionally, VLN tasks typically require agents to execute instructions in unseen environments, posing stringent tests on the model's generalization capability.

Currently, the mainstream paradigm for solving VLN tasks is imitation learning (IL), which employs DAgger-based~\cite{chen2022duet} techniques to train agents to mimic expert demonstrations through supervised learning. Although these methods have achieved significant results in seen environments, their optimization objective is to imitate expert behavior rather than learn goal-directed reasoning. Therefore, two fundamental limitations exist: (i) Limited generalization capability: IL agents rely on expert trajectories in training data, leading to overfitting and difficulty generalizing to unseen environments. (ii) Poor robustness to perturbations: IL methods suffer from the inherent distribution shift problem. When agents deviate from expert paths due to perturbations, they enter unseen state distributions. Without effective recovery policies, such deviations often prevent task completion.

Figure~\ref{fig:teaser} intuitively shows that in unseen environments, when IL agents are perturbed and deviate from expert paths, they often have to take significant detours due to a lack of effective recovery strategies. This raises the core question of this paper: \textit{Can we train a VLN agent that not only overcomes the limited generalization capability of IL, but also possesses effective recovery and replanning capabilities when perturbed and deviating from expert paths?} 

We hypothesize that robust learning requires exposing agents to \textbf{diverse navigation trajectories}, including both successes and failures. By learning relative distinctions across all outcomes, rather than relying solely on the absolute correctness of expert demonstrations, an agent can extract a richer learning signal that generalizes beyond simple expert replication.

To realize this hypothesis, we propose NavGRPO, a reinforcement learning framework designed for robust generalization built on three key design choices. (1) Group Relative Policy Optimization~\cite{guo2025deepseek}. GRPO samples multiple trajectories per instruction and optimizes policies through within-group performance comparisons. Unlike value-based RL methods~\cite{chen2021hamt} that require auxiliary critic networks, GRPO directly learns from trajectory-level rewards by ranking outcomes within each group. This design naturally encourages exploration of diverse navigation strategies while avoiding the instability of value network training, enabling agents to discover effective recovery behaviors beyond expert demonstrations. (2) Goal-oriented trajectory reward. Unlike prior RL methods using step-wise rewards~\cite{wang2019rcm}, our reward directly evaluates complete trajectories based on goal achievement and path efficiency, providing clearer learning signals. (3) Adaptive training strategy. We incorporate targeted supervised fine-tuning on challenging instructions to prevent catastrophic forgetting and complement reinforcement learning with behavioral guidance.

On R2R~\cite{anderson2018r2r} and REVERIE~\cite{qi2020reverie} validation unseen splits, we achieve +3\% and +1.71\% SPL improvements over the ScaleVLN baseline. More importantly, under extreme perturbations where agents are forced off-path in early steps, we demonstrate superior robustness with \textbf{+14.89\% SPL} improvement compared to ScaleVLN, showing effective robustness. Our contributions are: (1) We establish an effective framework for VLN that integrates GRPO with a multi-level reward function and adaptive training strategy, applicable to multiple baseline models. (2) Perturbation experiments show that agents trained with GRPO exhibit stronger robustness to perturbations present in navigation. (3) On standard R2R and REVERIE unseen benchmarks, we provide empirical evidence that goal-directed RL fine-tuning improves generalization to unseen environments.

%% file: sec/2_related_work.tex
\section{Related Work}
\label{sec:rel_work}

\noindent\textbf{Vision-and-Language Navigation (VLN).}
VLN requires embodied agents to follow natural language instructions to reach target locations in photo-realistic environments. The task was formalized with the Matterport3D simulator~\cite{chang2017matterport3d}, and various datasets have since been introduced to address different aspects of embodied navigation~\cite{anderson2018r2r,zhu2021soon,qi2020reverie,anderson2020rxr,jain2019stay,krantz2020beyond}, covering diverse scenarios from fine-grained object interaction to long-horizon instruction following across indoor and outdoor scenes.
Early approaches employed sequence-to-sequence architectures with imitation learning~\cite{anderson2018r2r,fried2018speaker,li2019robust}, which were later complemented by diverse training strategies including reinforcement learning~\cite{wang2019rcm,wang2018look,wang2020soft}, adversarial training~\cite{zhang2020language,fu2020counterfactual,lin2021adversarial}, generative modeling~\cite{kurita2020generative}, curriculum learning~\cite{zhang2021curriculum}, and cycle-consistent learning~\cite{wang2022counterfactual}. The integration of vision-language pre-training brought substantial improvements through large-scale offline pre-training~\cite{hao2020prevalent,huang2019transferable,hao2020towards,majumdar2020improving,lin2021sia,guhur2021airbert,hong2023learning,qiao2022hop,cui2023grounded}, auxiliary task design~\cite{ma2019self,zhu2020vision,li2023layout,wang2023lana,liang2022contrastive}, and regularization techniques~\cite{wang2020environment,parvaneh2020counterfactual,Wang2024GOAT} for more stable and less biased training. Architectural innovations evolved from recurrent encoders to sophisticated representations, including attention-based mechanisms~\cite{chen2021hamt,hong2020recurrent}, graph-based topological reasoning~\cite{deng2020evolving,wang2021structured,chen2021topological,chen2022duet}, and scene representations~\cite{anderson2019chasing,liu2024volumetric,chen2022sevol,lin2022multimodal,wang2023dual,an2023bevbert,liu2023bird,wang2023gridmm}, with recent works designing flexible action spaces for efficient exploration and backtracking~\cite{ke2019tactical,wang2020active,chen2022duet,hwang2023meta,gao2023adaptive}. To enhance cross-modal understanding, researchers have pursued finer-grained visual feature extraction~\cite{hu2019you,zhang2020diagnosing,qi2021road,moudgil2021soat,lin2023actional,huo2023geovln} and textual decomposition~\cite{zhu2020babywalk,qi2020object,an2021neighbor,hong2020graph,hong2020sub,li2021improving,zhang2023vlntrans,lin2022adapt,cheng2022learning}, while incorporating external knowledge from large language models~\cite{qiao2023march,qiao2024llm,zhou2023navgpt,chen2024mapgpt,zhou2024navgpt,lin2024correctable,zheng2024navillm,pan2023langnav,long2024discuss}, vision-language models~\cite{li2023layout}, and structured knowledge bases~\cite{li2023kerm}. Parallel efforts in data augmentation have explored observation perturbation~\cite{li2022envedit,koh2022simple,he2023frequency}, automatic trajectory annotation via speaker-follower frameworks~\cite{fried2018speaker,huang2019multi,tan2019envdrop,wang2022less,liang2022visual,kong2024controllable,wang2024bootstrapping}, and large-scale scene generation~\cite{liu2021envmixup,kamath2023new,chen2022hm3dlearning,wang2023scalevln,lin2023learning,li2024panogen}.
Despite these advances, existing methods predominantly rely on imitation learning from expert demonstrations, limiting exposure to diverse navigation scenarios. This results in poor generalization to novel configurations and fragility under execution deviations. We address these limitations with NavGRPO, a trajectory-level reinforcement learning approach that learns from diverse navigation experiences to develop robust policies.

\noindent\textbf{Reinforcement Learning for VLN.}
While imitation learning dominates VLN research, several efforts have explored reinforcement learning to address generalization challenges. Early work by Wang et al.~\cite{wang2019rcm} introduced Reinforced Cross-Modal Matching (RCM), which employs a matching critic to provide intrinsic rewards for vision-language alignment, combined with sparse task rewards through policy gradient optimization. Their method further incorporated Self-Imitation Learning (SIL) to exploit successful trajectories in unseen environments, demonstrating improved generalization on R2R. Subsequent approaches focused on reward engineering: SERL~\cite{wang2020soft} proposed soft expert reward learning to distill expert demonstrations into dense reward signals, avoiding manual reward shaping while maintaining strong supervision from human annotations. More recent methods have integrated RL into their training pipelines, with HAMT~\cite{chen2021hamt} applying policy gradient fine-tuning after pretraining, and SEvol~\cite{chen2022sevol} using RL to refine graph-based scene representations. However, these methods face critical limitations: step-level sparse rewards lead to severe credit assignment issues in long-horizon navigation, while learned value networks struggle in VLN's high-dimensional action space. We adopt Group Relative Policy Optimization (GRPO)~\cite{guo2025deepseek}, which eliminates both issues through trajectory-level geometric rewards and advantage estimation via relative ranking, enabling robust learning from diverse trajectories including both successful and failed attempts.

%% file: sec/3_method.tex
\section{Method}

\begin{figure*}[t]
 \centering
\includegraphics[width=\textwidth]{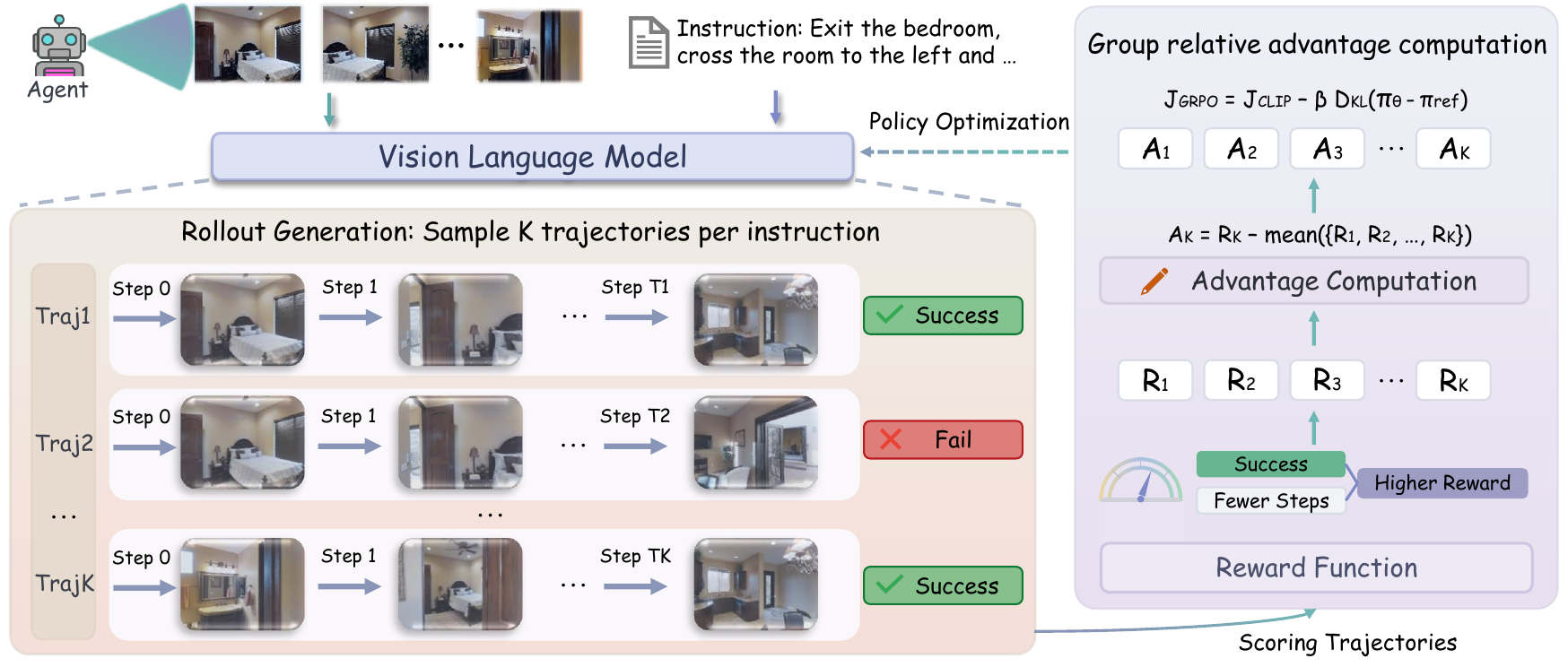}
\caption{Overview of our NavGRPO training framework for vision-language navigation. For each instruction, we sample K diverse trajectories through policy rollout, compute rewards using trajectory-level and step-level signals, estimate group relative advantages by comparing within instruction groups, and optimize the policy through debiased advantage estimation without value networks.}
\label{fig:method}
\end{figure*}

\noindent\textbf{Problem Definition.}
In the standard VLN setup, the environment is represented as an undirected navigation graph $\mathcal{G} = \{\mathcal{V}, \mathcal{E}\}$, where $\mathcal{V} = \{V_i\}_{i=1}^{K}$ denotes a set of $K$ navigable locations and $\mathcal{E}$ represents the connectivity edges between locations. Each location in the graph corresponds to a specific viewpoint in the physical environment, and the agent can only traverse along the edges defined by the graph structure. Given a natural language instruction $\mathcal{W} = \{w_1, w_2, ..., w_L\}$ composed of $L$ words, at each navigation timestep $t$, the agent is situated at location $V_t \in \mathcal{V}$ and perceives the surrounding environment through panoramic visual observations $\mathcal{O}_t = \{o_i\}_{i=1}^{n}$ captured at the current location, consisting of $n$ view images. The agent also observes a set of reachable neighboring nodes $\mathcal{N}(V_t)$. Based on the instruction and current observations, the agent must select an action $a_t$ from the available action space $\mathcal{A}_t$. The action space includes navigating to a neighboring location $V_{t+1} \in \mathcal{N}(V_t)$ connected to the current position, or issuing a stop signal to terminate the navigation episode.

\noindent\textbf{Overview.}
NavGRPO trains navigation policies through Group Relative Policy Optimization, which learns from diverse trajectories sampled during training. We describe the GRPO framework in Sec.~\ref{sec:grpo}, our geometric reward design in Sec.~\ref{sec:reward}, and the training pipeline in Sec.~\ref{sec:train}.

\subsection{NavGRPO for VLN}
\label{sec:grpo}

We adopt GRPO~\cite{guo2025deepseek}, which estimates advantages through group-based reward comparison without requiring a separate value network.

\noindent\textbf{Policy Architecture.}
The proposed navigation policy $\pi_\theta$ is parameterized by the vision-language transformer network. At each timestep $t$, given the instruction $\mathcal{W}$, the current panoramic observations $\mathcal{O}_t$, and the graph map representation $\mathcal{M}_t$ that encodes spatial topology and historical navigation information, the policy outputs a probability distribution over the action space:
\begin{equation}
\pi_\theta(a_t | \mathcal{W}, \mathcal{O}_t, \mathcal{M}_t) = \text{softmax}(f_\theta(\mathcal{W}, \mathcal{O}_t, \mathcal{M}_t))
\end{equation}
where $f_\theta$ represents the network's logit output for each candidate action in $\mathcal{A}_t$.

\noindent\textbf{Group-based Trajectory Sampling.}
For each instruction $\mathcal{W}_i$ sampled from the instruction set $\mathcal{D}_{\mathcal{B}}$, which denotes a mini-batch of $B$ instructions, we sample a group of $K$ independent trajectories by rolling out the current policy $\pi_\theta$. Each trajectory $\tau_k = (s_{k,0}, a_{k,0}, s_{k,1}, a_{k,1}, ..., s_{k,T})$ represents a complete navigation episode, where $s_{k,t} = (V_{k,t}, \mathcal{O}_{k,t}, \mathcal{M}_{k,t})$ denotes the state at timestep $t$. During sampling, actions are drawn stochastically from $a_t \sim \pi_\theta(\cdot | s_{k,t})$, and we store the corresponding log probability $\log \pi_\theta(a_{k,t} | s_{k,t})$ as $p_{k,t}^{\text{old}}$ for later policy updates.

\noindent\textbf{Debiased Group Relative Advantage Estimation.}
For each instruction $\mathcal{W}$ in the training batch, we sample $K$ trajectories and compute their rewards $\{r_1, r_2, ..., r_K\}$ using the reward function detailed in Section~\ref{sec:reward}. Following Dr.GRPO~\cite{liu2025understanding}, we estimate advantages through group-relative comparison without variance normalization:
\begin{equation}
\hat{A}_{k} = r_k - \text{mean}(\{r_1, ..., r_K\})
\end{equation}
This uses the empirical group mean as a natural baseline, eliminating the need for separate value function approximation. Removing variance normalization avoids amplifying noise in low-diversity groups, making the learning signal more robust. The trajectory-level advantage is broadcast to all timesteps during policy updates.

\noindent\textbf{Policy Optimization Objective.}
For each state-action transition $(s_{k,t}, a_{k,t})$ in trajectory $\tau_k$, we compute the probability ratio:
\begin{equation}
\rho_{k,t} = \frac{\pi_\theta(a_{k,t} | s_{k,t})}{\pi_{\theta_{\text{old}}}(a_{k,t} | s_{k,t})} = \exp\left(\log \pi_\theta(a_{k,t} | s_{k,t}) - p_{k,t}^{\text{old}}\right)
\end{equation}
where $\pi_{\theta_{\text{old}}}$ represents the behavior policy with frozen parameters. We incorporate the step-level progress coefficient $\gamma_{k,t}$ from Section~\ref{sec:reward} to modulate the advantage signal. Following PPO~\cite{schulman2017proximal}, the clipped surrogate objective becomes:
\begin{equation}
\mathcal{J}_{\text{clip}}(k,t) = \min\Big(\gamma_{k,t} \hat{A}_{k} \rho_{k,t}, \; \gamma_{k,t} \hat{A}_{k} \cdot \text{clip}(\rho_{k,t}, 1\!-\!\delta, 1\!+\!\delta)\Big)
\end{equation}
The clipping mechanism constrains $\rho_{k,t}$ within $[1-\delta, 1+\delta]$ to prevent excessively large policy updates. The complete optimization objective is:
\begin{equation}
\begin{aligned}
\mathcal{J}_{\text{GRPO}}(\theta) &= \mathbb{E}_{\mathcal{W} \sim \mathcal{D}_{\mathcal{B}}, \{\tau_k\}_{k=1}^K \sim \pi_{\theta_{\text{old}}}(\cdot | \mathcal{W})} \\
&\left[\frac{1}{K} \sum_{k=1}^{K} \sum_{t=1}^{|\tau_k|} \left(\mathcal{J}_{\text{clip}}(k,t) - \beta \mathbb{D}_{\text{KL}}[\pi_\theta \| \pi_{\text{ref}}]\right)\right]
\end{aligned}
\end{equation}
We aggregate across all timesteps without length normalization to avoid introducing bias. The KL divergence term regularizes the policy against the reference policy $\pi_{\text{ref}}$, which is the fixed policy before RL training begins, weighted by $\beta$ to prevent excessive deviation.

\input{tables/robust}

\subsection{Reward Function for NavGRPO}
\label{sec:reward}

The reward function guides the RL training process by providing learning signals at both trajectory and step levels. We design a composite reward function with trajectory-level rewards for overall navigation quality and step-level rewards for step-wise progress.

\noindent\textbf{Navigation Success Reward.}
Let $V_T^k$ denote the final viewpoint of trajectory $\tau_k$ and $V^*$ represent the target destination. The navigation error is $d_k = d_{\text{shortest}}(V_T^k, V^*)$. The success reward uses exponential decay:
\begin{equation}
R_{\text{nav}}(d_k) = \exp\left(-\frac{d_k^2}{2\epsilon^2}\right) \cdot \mathbb{I}(d_k < \epsilon)
\end{equation}
Where $\epsilon$ is the success threshold. This provides smooth decay that emphasizes precise goal reaching while maintaining differentiability.

\noindent\textbf{Path Efficiency Reward.}
To discourage inefficient behaviors such as backtracking and detours, we penalize excessive path length:
\begin{equation}
R_{\text{path}}(L_k, L^*) = -\max(L_k - L^*, 0) / L^*
\end{equation}
where $L_k$ is the actual path length and $L^*$ is the shortest path length.

\noindent\textbf{Total Reward Function.}
The trajectory-level reward combines navigation success and path efficiency:
\begin{equation}
r_k = R_{\text{nav}}(d_k) + \alpha \cdot R_{\text{path}}(L_k, L^*)
\end{equation}
where $\alpha$ balances the contribution of path efficiency.

\noindent\textbf{Step-level Progress Coefficient.}
Navigation tasks possess well-defined spatial metrics that allow quantifying progress at each decision step. For trajectory $\tau_k$ at timestep $t$, we define a progress coefficient that modulates the advantage signal:
\begin{equation}
\gamma_{k,t} = 1 + \text{sign}(\hat{A}_k) \cdot \frac{d_{t-1} - d_t}{L^*}
\end{equation}
where $d_t$ is the distance to the goal at step $t$ and $L^*$ is the 
ground-truth shortest path length. This coefficient ensures that steps 
approaching the goal ($d_{t-1} > d_t$) yield stronger learning signals 
$\gamma_{k,t} \cdot \hat{A}_k$ in magnitude, while steps deviating from 
the goal receive attenuated signals, allowing the agent to distinguish 
productive actions even in failed trajectories.

\subsection{Adaptive Training with Hard Case Replay}
\label{sec:train}

While our method enables learning from diverse trajectories, certain instructions remain persistently challenging despite extensive sampling. We address this by periodically applying supervised refinement on hard cases identified during RL training.

\noindent\textbf{Hard Case Identification.}
For each instruction $\mathcal{W}_i$ in the training batch, we sample $K$ trajectories and identify it as a hard case when all sampled trajectories fail:
\begin{equation}
\text{Hard}(\mathcal{W}_i) = \mathbb{I}\left(\sum_{k=1}^{K} \mathbb{I}(d_k < \epsilon) = 0\right)
\end{equation}
These instructions are stored in buffer $\mathcal{B}_{\text{hard}}$.

\noindent\textbf{Supervised Refinement.}
When $|\mathcal{B}_{\text{hard}}|$ reaches threshold $M$, we perform supervised updates on expert trajectories:
\begin{equation}
\mathcal{L}_{\text{hard}}(\theta) = -\frac{1}{|\mathcal{B}_{\text{hard}}|} \sum_{\mathcal{W}_i \in \mathcal{B}_{\text{hard}}} \sum_{t=0}^{T_i} \log \pi_\theta(a_t^* | s_t^*, \mathcal{W}_i)
\end{equation}
The buffer is then cleared. This adds negligible cost since the compute saved from fewer RL rollouts more than compensates for the supervised updates on hard cases.

%% file: tables/robust.tex

\begin{table*}[t]
\centering
\caption{Robustness analysis under stochastic perturbations on R2R Val Unseen. Left: Global perturbation samples from policy distribution with probability $p$. Right: Early perturbation selects the least probable action for the first $N$ steps. $\Delta$SPL shows SPL degradation from the unperturbed setting (prob=0 or Steps=0) for each method. NavGRPO demonstrates superior robustness across all perturbation levels.}
\label{tab:robustness}
\renewcommand{\arraystretch}{1.2}
\small
\setlength{\tabcolsep}{8pt}
\begin{tabular}{l c ccccc c ccccc}
\hline
\multirow{2}{*}{Method} & \multicolumn{6}{c}{Global Perturbation} & \multicolumn{6}{c}{Early Perturbation} \\
\cline{2-7} \cline{8-13}
& prob & OSR$\uparrow$ & NE$\downarrow$ & SR$\uparrow$ & SPL$\uparrow$ & $\Delta$SPL & Steps & OSR$\uparrow$ & NE$\downarrow$ & SR$\uparrow$ & SPL$\uparrow$ & $\Delta$SPL \\
\hline
ScaleVLN & 0.0 & 87.48 & 2.34 & 79.40 & 69.97 & -0.00 & 0 & 87.48 & 2.34 & 79.40 & 69.97 & -0.00 \\
\rowcolor{gray!20} NavGRPO & 0.0 & \textbf{89.18} & \textbf{2.19} & \textbf{81.88} & \textbf{72.65} & \textbf{-0.00} & 0 & \textbf{89.18} & \textbf{2.19} & \textbf{81.88} & \textbf{72.65} & \textbf{-0.00} \\
\hline
ScaleVLN & 0.2 & 87.14 & 2.43 & 78.97 & 67.80 & -2.17 & 1 & 83.06 & 2.94 & 74.76 & 59.06 & -10.91 \\
\rowcolor{gray!20} NavGRPO & 0.2 & \textbf{88.75} & \textbf{2.21} & \textbf{81.27} & \textbf{72.11} & \textbf{-0.54} & 1 & \textbf{85.05} & \textbf{2.67} & \textbf{77.44} & \textbf{67.04} & \textbf{-5.61} \\
\hline
ScaleVLN & 0.4 & 87.19 & 2.42 & 78.67 & 65.90 & -4.07 & 2 & 83.61 & 2.86 & 75.69 & 54.25 & -15.72 \\
\rowcolor{gray!20} NavGRPO & 0.4 & \textbf{88.79} & \textbf{2.22} & \textbf{81.06} & \textbf{71.69} & \textbf{-0.96} & 2 & \textbf{83.85} & \textbf{2.70} & \textbf{76.50} & \textbf{65.10} & \textbf{-7.55} \\
\hline
ScaleVLN & 0.8 & 87.06 & 2.52 & 77.61 & 61.94 & -8.03 & 3 & 83.52 & 3.04 & 74.16 & 47.77 & -22.20 \\
\rowcolor{gray!20} NavGRPO & 0.8 & \textbf{88.49} & \textbf{2.27} & \textbf{79.95} & \textbf{69.98} & \textbf{-2.67} & 3 & \textbf{82.96} & \textbf{2.82} & \textbf{75.35} & \textbf{62.66} & \textbf{-9.99} \\
\hline
\end{tabular}
\end{table*}

%% file: sec/4_experiment.tex
\section{Experiment}
\subsection{Experimental Setup}

\input{tables/overall}
\input{tables/r2r-ce}

\noindent\textbf{Datasets.}
We evaluate on three widely-used VLN benchmarks: R2R~\cite{anderson2018r2r} provides fine-grained step-by-step navigation instructions; REVERIE~\cite{qi2020reverie} presents high-level goal-oriented instructions specifying target rooms and objects; R2R-CE~\cite{krantz2020beyond} extends R2R to continuous environments using Habitat simulator~\cite{savva2019habitat}, where agents navigate continuously rather than between discrete viewpoints.

\noindent\textbf{Evaluation Metrics.}
We adopt standard VLN evaluation metrics~\cite{anderson2018r2r}. Navigation Error (NE) measures the average distance in meters between the agent's final position and the goal location. Success Rate (SR) computes the percentage of episodes where the agent stops within 3 meters of the target. Success Rate penalized by Path Length (SPL) penalizes SR by the ratio of the shortest path length to the actual trajectory length, rewarding both accuracy and efficiency. Oracle Success Rate (OSR) measures the success rate under an ideal stop policy.

\noindent\textbf{Implementation Details.}
We maintain the same setup as baseline methods~\cite{chen2022duet,wang2023scalevln}. All models are trained for 200k steps with learning rate $1\times10^{-5}$. After supervised warm-up (30k steps), we transition to GRPO with batch size $B=8$ and $K=8$ trajectories per instruction. Following standard PPO settings~\cite{schulman2017proximal}, we set clipping threshold $\delta=0.2$ and KL penalty $\beta=0.01$. For reward function, we use $\alpha=0.25$. For hard case replay, we set buffer size M = 200. As training progresses and the policy improves, the frequency of hard case updates naturally decreases. All results are averaged over three random seeds.

\subsection{Robustness to Action Noise}

\noindent\textbf{Motivation.}
Real-world deployment faces execution noise from sensor errors and actuation imprecision. While imitation learning optimizes for near-optimal trajectories, it provides limited exposure to noisy action sequences during training. We evaluate whether RL exploration, by experiencing diverse state-action distributions, enables the agent to better handle stochastic perturbations at inference time.

\noindent\textbf{Experimental Setup.}
We design two perturbation strategies to evaluate robustness: (1) \textit{Global perturbation} mimics real-world random noise—at each step with probability $p \in \{0.0, 0.2, 0.4, 0.8\}$, the agent samples an action from its policy distribution $\pi(\cdot|s)$; otherwise it takes the argmax action. (2) \textit{Early perturbation} tests recovery from initial mistakes—the first $N \in \{1, 2, 3\}$ steps select the \textit{least probable} action from the policy distribution, while remaining steps use argmax actions. This simulates scenarios where agents start from poor decisions due to uncertain states, such as imprecise localization, but must recover to reach the goal.

\noindent\textbf{Results and Analysis.}
Table~\ref{tab:robustness} demonstrates superior robustness of our method across both perturbation scenarios. Under global perturbation at $p=0.4$, our method degrades by only 0.96 SPL compared to the baseline's 4.07 degradation, representing 4.2$\times$ better robustness. At $p=0.8$, this advantage amplifies to 3.0$\times$ with degradations of 2.67 versus 8.03 SPL. Notably, our method maintains 69.98 SPL at $p=0.8$, exceeding the baseline's unperturbed performance of 69.97. For early perturbation, perturbing the first three steps causes our method to degrade by 9.99 SPL while the baseline degrades by 22.20 SPL, yielding a 2.2$\times$ robustness advantage. These results indicate that GRPO training enables effective recovery from execution perturbations through exposure to diverse trajectories.

\subsection{Main Results}

\noindent\textbf{R2R.} 
Table~\ref{tab:main_results} shows results on the R2R dataset. Our method consistently outperforms DUET, achieving 2\% higher SR and 3\% higher SPL on both val unseen and test unseen splits. When applied to ScaleVLN, our approach achieves 3\% SPL improvement over the baseline and outperforms GOAT by 5\% on val unseen.

\noindent\textbf{REVERIE.} 
On the REVERIE dataset, which features high-level goal-oriented instructions requiring longer-horizon planning, our approach outperforms DUET by 2.49\% in SR and 1.59\% in SPL on val unseen. When built upon ScaleVLN, our method further improves SR by 1.94\% and SPL by 1.71\%, surpassing GOAT by 6.48\% in SPL. The gains are validated on test unseen, where we achieve 1.98\% higher SR and 2.10\% higher SPL compared to DUET. Notably, our approach substantially outperforms traditional RL-based methods such as RCM and HAMT, which face challenges from step-level sparse rewards and credit assignment in long-horizon navigation. The consistent improvements across both datasets demonstrate generalization to diverse navigation scenarios, from fine-grained step-by-step instructions to high-level goal-oriented tasks.

\noindent\textbf{R2R-CE.}
Table~\ref{tab:r2r-ce} presents results on the continuous R2R-CE benchmark. Despite training in discrete environments, both DUET-GRPO and ScaleVLN-GRPO show consistent improvements over their respective baselines, demonstrating effective transfer to continuous navigation scenarios.

\subsection{Ablation Studies and Analysis}

\input{tables/reward}
\input{tables/grpo_variants}

\noindent\textbf{Impact of Reward Function Design.}
To understand the contribution of different reward components, we conduct
ablation studies on the R2R val unseen split, as shown in Table~\ref{tab:reward_ablation}. Using only the navigation success reward provides a basic foundation with 71.07\% SPL, demonstrating that sparse trajectory-level signals alone can guide policy learning. Incorporating the path efficiency reward improves SPL to 71.88\%, indicating that penalizing unnecessarily long trajectories encourages more compact navigation behaviors. Alternatively, adding the step-level progress reward achieves 71.93\% SPL, showing that fine-grained intermediate guidance helps the agent make better local decisions. Combining all three components yields the best performance at 72.65\% SPL and 81.88\% SR, demonstrating that trajectory-level and step-level rewards provide complementary supervision—the former guides overall navigation quality while the latter refines individual action selections.

\noindent\textbf{Impact of Group-Based Policy Optimization.}
We compare different variants of group-based policy optimization to validate our design choices, as shown in Table~\ref{tab:grpo_variants}. Without grouping, the agent achieves 69.97\% SPL on val unseen, establishing a baseline for individual trajectory optimization. Standard GRPO~\cite{guo2025deepseek} introduces group-wise advantage normalization, improving performance to 71.23\% SPL through better calibrated gradients. We evaluate three advanced variants adapted from recent LLM alignment literature: GSPO~\cite{zheng2025group} shifts importance sampling and clipping to the sequence level, GMPO~\cite{zhao2025geometric} adopts geometric mean for step-level reward aggregation, and Dr.GRPO~\cite{liu2025understanding} removes length and variance normalization to mitigate length bias. Dr.GRPO achieves the strongest performance with 72.65\% SPL on val unseen. We adopt Dr.GRPO's debiased advantage estimation in our framework, which removes normalization constraints and reduces hyperparameter sensitivity. This design choice reduces hyperparameter sensitivity and proves effective for generalizing across different VLN benchmarks and base models. The consistent gains across different optimization variants confirm the robustness of this approach.

\input{tables/trajectory_number}

\noindent\textbf{Analysis of Sampling Trajectory Number.}
We investigate the impact of sampling trajectory number $K$ during training, where $K$ trajectories are sampled for each instruction to form a group for relative advantage computation. Table~\ref{tab:trajectory_number} shows the results on R2R val unseen split with different values of $K$. Using fewer trajectories provides limited diversity for relative comparison, resulting in suboptimal performance. Increasing $K$ to 8 substantially improves both SR and SPL by 1.51\% and 1.63\% respectively compared to $K=4$, as the agent benefits from richer comparative signals within each group. Further increasing $K$ to 16 yields only marginal gains of 0.08\% in SR and 0.22\% in SPL while doubling the training time and memory cost. Therefore, we adopt $K=8$ as our default setting to balance performance and computational efficiency.

\input{tables/RL_comparison}
\input{tables/training_strategy}

\noindent\textbf{Comparison with Alternative RL Methods.}
We apply different RL algorithms to further optimize the IL-finetuned model in Table~\ref{tab:rl_methods} under a unified training framework. Our GRPO uses trajectory-level rewards measuring navigation success and path efficiency, which cannot be decomposed into step-level signals required for A2C and PPO's value bootstrapping. REINFORCE, though compatible with both reward types, still fails to improve performance due to its inherently high gradient variance. Therefore, following established practices~\cite{chen2021hamt}, we provide classical methods with step-level rewards including distance progress and orientation alignment. REINFORCE degrades performance slightly, while A2C achieves modest gains of 0.28\% SPL, and PPO improves the SPL metric to 70.68\%. In contrast, our GRPO achieves 72.65\% SPL, outperforming PPO by a margin of 1.97\%. This result demonstrates that comparing trajectories within instruction groups yields substantially more informative learning signals than optimizing trajectories independently.

\begin{figure}[ht]
 \centering
\includegraphics[width=\linewidth]{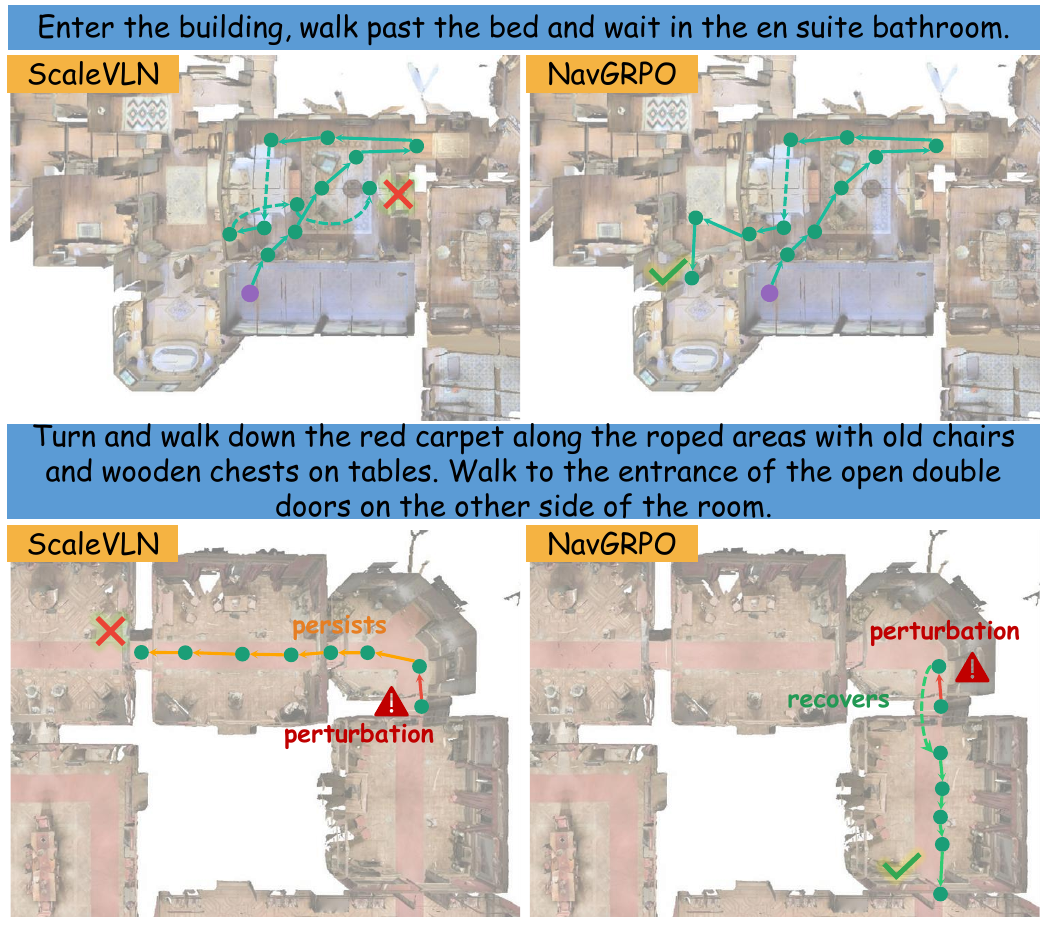}
\caption{Qualitative comparison on challenging instructions under normal conditions (top) and initial perturbations (bottom). ScaleVLN fails to recover from errors in both scenarios. Our GRPO-trained agent successfully completes tasks and demonstrates robust error correction under perturbations.}
\label{fig:result}
\end{figure}

\noindent\textbf{Analysis of Training Strategy.}
Table~\ref{tab:training_strategy} compares different training strategies to understand the contribution of each component. All methods start from the same pretrained vision-language model. Training with RL alone performs poorly without navigation-specific initialization. SFT-only training establishes a solid baseline by learning from expert demonstrations. Sequential SFT-RL improves performance to 81.25\% SR and 72.08\% SPL, demonstrating that RL optimization can surpass IL-finetuned models. Our approach with hard case replay further enhances results to 81.88\% SR and 72.65\% SPL, providing modest but consistent gains. Importantly, hard case replay is triggered only when the entire sampled group fails, thereby avoiding redundant RL exploration on persistently challenging instructions. This adaptive strategy stabilizes training and prevents catastrophic forgetting, balancing broad policy exploration with targeted supervision.

\subsection{Qualitative Analysis}
Figure~\ref{fig:result} presents qualitative comparisons on spatially ambiguous instructions. Under standard conditions, ScaleVLN commits navigation errors from which it cannot recover, while our GRPO-trained agent makes correct decisions at critical waypoints to reach target locations. Under adversarial perturbations, ScaleVLN persists along erroneous trajectories initiated by the perturbation. In contrast, our method demonstrates error-correction capabilities: in the first case, it recognizes spatial deviation and backtracks to the correct path; in the second case, it adjusts mid-trajectory despite initial disruption. This demonstrates that group-based trajectory optimization enables both improved spatial reasoning and robust recovery from navigational errors.

\noindent \textbf{Additional analysis, including more RL comparisons, hyperparameter sensitivity, computational cost, visualizations, and limitations, is provided in the Appendix.}

%% file: tables/overall.tex
\begin{table*}[t]
\centering
\caption{Comparison with SoTA methods on R2R and REVERIE datasets. $\dagger$: Methods using reinforcement learning for policy optimization.}
\label{tab:main_results}
\renewcommand{\arraystretch}{1}
\small
\setlength{\tabcolsep}{7.1pt}
\begin{tabular}{l ccc ccc ccc ccc}
\hline
\multirow{2}{*}{Methods} & \multicolumn{3}{c}{R2R Val Unseen} & \multicolumn{3}{c}{R2R Test Unseen} & \multicolumn{3}{c}{REVERIE Val Unseen} & \multicolumn{3}{c}{REVERIE Test Unseen} \\
\cline{2-13}
& NE$\downarrow$ & SR$\uparrow$ & SPL$\uparrow$ & NE$\downarrow$ & SR$\uparrow$ & SPL$\uparrow$ & OSR$\uparrow$ & SR$\uparrow$ & SPL$\uparrow$ & OSR$\uparrow$ & SR$\uparrow$ & SPL$\uparrow$ \\
\hline
RCM$^{\dagger}$~\cite{wang2019rcm} & 5.88 & 43 & - & 6.12 & 43 & 38 & 14.23 & 9.29 & 6.97 & 11.68 & 7.84 & 6.67 \\
SERL$^{\dagger}$~\cite{wang2020soft} & 4.74 & 56 & 48 & 5.63 & 53 & 49 & - & - & - & - & - & - \\
VLN$\circlearrowright$BERT~\cite{hong2020recurrent} & 3.93 & 63 & 57 & 4.09 & 63 & 57 & 35.02 & 30.67 & 24.90 & 32.91 & 29.61 & 23.99 \\
HAMT$^{\dagger}$~\cite{chen2021hamt} & 3.65 & 66 & 61 & 3.93 & 65 & 60 & 36.84 & 32.95 & 30.20 & 33.41 & 30.40 & 26.67 \\
SEvol$^{\dagger}$~\cite{chen2022sevol} & 3.99 & 62 & 57 & 4.13 & 62 & 57 & - & - & - & - & - & - \\
HOP+~\cite{qiao2023hop+} & 3.49 & 67 & 61 & 3.71 & 66 & 60 & 40.04 & 36.07 & 31.13 & 35.81 & 33.82 & 28.24 \\
BEVBert~\cite{an2023bevbert} & 2.81 & 75 & 64 & 3.13 & 73 & 62 & 56.40 & 51.78 & 36.37 & 57.26 & 52.81 & 36.41 \\
LANA~\cite{wang2023lana} & - & 68 & 62 & - & 65 & 60 & 38.54 & 34.00 & 29.26 & 36.41 & 33.50 & 26.89 \\
KERM~\cite{li2023kerm} & 3.22 & 72 & 61 & 3.61 & 70 & 59 & 55.21 & 50.44 & 35.38 & 57.58 & 52.43 & 39.21 \\
GridMM~\cite{wang2023gridmm} & 2.83 & 75 & 64 & 3.35 & 73 & 62 & 57.48 & 51.37 & 36.47 & 59.55 & 53.13 & 36.60 \\
NaviLLM~\cite{zheng2024navillm} & 3.51 & 67 & 59 & 3.71 & 68 & 60 & 52.27 & 42.15 & 35.68 & 51.75 & 39.80 & 32.33 \\
NavGPT-2~\cite{zhou2024navgpt} & 2.84 & 74 & 61 & 3.33 & 72 & 60 & - & - & - & - & - & - \\
VER~\cite{liu2024volumetric} & 2.80 & 76 & 65 & 2.74 & 76 & 66 & 61.09 & 55.98 & 39.66 & 62.22 & 56.82 & 38.76 \\
MAGIC-L~\cite{wang2024magic} & 2.22 & 79 & 70 & 2.75 & 77 & 69 & - & - & - & - & - & - \\
GOAT~\cite{Wang2024GOAT} & 2.40 & 78 & 68 & 3.04 & 75 & 65 & - & 53.37 & 36.70 & - & 57.72 & 40.53 \\
NavQ~\cite{xu2025navq} & 3.06 & 73 & 63 & 3.30 & 72 & 63 & 60.47 & 53.22 & 38.89 & 60.39 & 53.29 & 39.50 \\
COSMO~\cite{zhang2025cosmo} & 3.15 & 73 & 61 & 3.43 & 71 & 58 & 56.09 & 50.81 & 35.93 & 59.33 & 52.53 & 36.12 \\

\hline
DUET~\cite{chen2022duet} & 3.31 & 72 & 60 & 3.65 & 69 & 59 & 51.07 & 46.98 & 33.73 & 56.91 & 52.51 & 36.06 \\
\rowcolor{gray!20} DUET-NavGRPO & 3.18 & 74 & 63 & 3.39 & 71 & 62 & 53.17 & 49.47 & 35.32 & 58.32 & 54.49 & 38.16 \\
\hline
ScaleVLN~\cite{wang2023scalevln} & 2.34 & 79 & 70 & 2.73 & 77 & 68 & 63.85 & 56.97 & 41.84 & 62.65 & 56.13 & 39.52 \\
\rowcolor{gray!20} ScaleVLN-NavGRPO & \textbf{2.19} & \textbf{82} & \textbf{73} & \textbf{2.52} & \textbf{79} & \textbf{70} & \textbf{65.19} & \textbf{58.91} & \textbf{43.55} & \textbf{64.21} & \textbf{58.25} & \textbf{41.34} \\
\hline
\end{tabular}
\end{table*}

%% file: tables/r2r-ce.tex
\begin{table}[t]
\centering
\caption{Navigation performance on the R2R-CE dataset. $\dagger$: Methods that apply candidate waypoint predictor to support high-level action space.}
\label{tab:r2r-ce}
\small
\setlength{\tabcolsep}{3.4pt}
\begin{tabular}{l ccc ccc}
\hline
\multirow{2}{*}{Methods} & \multicolumn{3}{c}{Val Unseen} & \multicolumn{3}{c}{Test Unseen} \\
\cline{2-7}
& NE$\downarrow$ & SR$\uparrow$ & SPL$\uparrow$ & NE$\downarrow$ & SR$\uparrow$ & SPL$\uparrow$ \\
\hline
LAW~\cite{raychaudhuri2021law} & 6.83 & 35 & 31 & - & - & - \\
Sim2Sim~\cite{krantz2022sim} & 6.07 & 43 & 36 & 6.17 & 44 & 37 \\
MGMap~\cite{chen2022weakly} & 6.28 & 39 & 34 & 7.11 & 35 & 28 \\
CMA$\dagger$~\cite{hong2022bridging} & 6.20 & 41 & 36 & 6.30 & 38 & 33 \\
VLN$\circlearrowright$BERT$\dagger$~\cite{hong2022bridging} & 5.74 & 44 & 39 & 5.89 & 42 & 36 \\
GridMM$\dagger$~\cite{wang2023gridmm} & 5.11 & 49 & 41 & 5.64 & 46 & 39 \\
Ego$^2$-Map$\dagger$ & 4.94 & 52 & 46 & 5.54 & 47 & 41 \\
Reborn~\cite{an20221st} & 5.40 & 50 & 46 & 5.55 & 49 & 45 \\
ETPNAV~\cite{an2024etpnav} & 4.71 & 57 & 49 & 5.12 & 55 & 48 \\
COSMO$\dagger$~\cite{zhang2025cosmo} & - & 47 & 40 & - & 47 & 40 \\
\hline
DUET$\dagger$~\cite{chen2022duet} & 5.26 & 47 & 39 & 5.82 & 42 & 36 \\
\rowcolor{gray!20} DUET-NavGRPO$\dagger$ & 5.02 & 50 & 42 & 5.64 & 44 & 38 \\
\hline
ScaleVLN$\dagger$ & 4.80 & 55 & 51 & 5.11 & 55 & 50 \\
\rowcolor{gray!20} ScaleVLN-NavGRPO$\dagger$ & \textbf{4.69} & \textbf{57} & \textbf{53} & \textbf{5.01} & \textbf{57} & \textbf{52} \\
\hline
\end{tabular}
\end{table}

%% file: tables/reward.tex
\begin{table}[t]
\centering
\caption{Ablation study on reward function design. Experiments are conducted on R2R val unseen split.}
\label{tab:reward_ablation}
\small
\setlength{\tabcolsep}{6.4pt}
\begin{tabular}{c ccc ccc}
\hline
\multirow{2}{*}{Method \#} & \multicolumn{3}{c}{Reward Components} & \multicolumn{3}{c}{R2R Val Unseen} \\
\cline{2-7}
& R$_{\text{nav}}$ & R$_{\text{path}}$ & R$_{\text{step}}$ & NE$\downarrow$ & SR$\uparrow$ & SPL$\uparrow$ \\
\hline
1 & \checkmark & & & 2.31 & 80.14 & 71.07 \\
2 & \checkmark & \checkmark & & 2.26 & 81.02 & 71.88 \\
3 & \checkmark & & \checkmark & 2.25 & 81.29 & 71.93 \\
4 & \checkmark & \checkmark & \checkmark & \textbf{2.19} & \textbf{81.88} & \textbf{72.65} \\
\hline
\end{tabular}
\end{table}

%% file: tables/grpo_variants.tex
\begin{table}[t]
\centering
\caption{Comparison of group-based policy optimization variants on R2R Val Unseen split. All methods use the same reward function and sampling strategy.}
\label{tab:grpo_variants}
\small
\setlength{\tabcolsep}{9.5pt}
\begin{tabular}{l cccc}
\hline
Method & OSR$\uparrow$ & NE$\downarrow$ & SR$\uparrow$ & SPL$\uparrow$ \\
\hline
w/o Group & 87.48 & 2.34 & 79.40 & 69.97 \\
GRPO~\cite{guo2025deepseek} & 88.15 & 2.26 & 80.15 & 71.23 \\
GSPO~\cite{zheng2025group} & 88.62 & 2.23 & 80.52 & 71.68 \\
GMPO~\cite{zhao2025geometric} & 88.56 & 2.24 & 80.47 & 71.62 \\
Dr.GRPO~\cite{liu2025understanding} & \textbf{89.02} & \textbf{2.19} & \textbf{81.88} & \textbf{72.65} \\
\hline
\end{tabular}
\end{table}

%% file: tables/trajectory_number.tex
\begin{table}[t]
\centering
\caption{Analysis of sampling trajectory number on R2R Val Unseen split.}
\label{tab:trajectory_number}
\small
\setlength{\tabcolsep}{9.5pt}
\begin{tabular}{l cccc}
\hline
Trajectories & OSR$\uparrow$ & NE$\downarrow$ & SR$\uparrow$ & SPL$\uparrow$ \\
\hline
$K=2$ & 88.28 & 2.31 & 79.98 & 71.02 \\
$K=4$ & 88.62 & 2.28 & 80.37 & 71.98 \\
$K=8$ & 89.02 & 2.19 & 81.88 & 72.65 \\
$K=16$ & \textbf{89.12} & \textbf{2.15} & \textbf{81.96} & \textbf{72.87} \\
\hline
\end{tabular}
\end{table}

%% file: tables/RL_comparison.tex
\begin{table}[t]
\centering
\caption{Comparison with alternative RL methods on R2R Val Unseen split.}
\label{tab:rl_methods}
\small
\setlength{\tabcolsep}{8pt}
\begin{tabular}{l cccc}
\hline
Method & OSR$\uparrow$ & NE$\downarrow$ & SR$\uparrow$ & SPL$\uparrow$ \\
\hline
SFT only & 87.48 & 2.34 & 79.40 & 69.97 \\
SFT+REINFORCE & 87.35 & 2.36 & 79.28 & 69.85 \\
SFT+A2C & 87.72 & 2.32 & 79.65 & 70.25 \\
SFT+PPO & 88.15 & 2.29 & 79.95 & 70.68 \\
SFT+GRPO & \textbf{89.02} & \textbf{2.19} & \textbf{81.88} & \textbf{72.65} \\
\hline
\end{tabular}
\end{table}

%% file: tables/training_strategy.tex
\begin{table}[t]
\centering
\caption{Analysis of different training strategies on R2R Val Unseen split.}
\label{tab:training_strategy}
\small
\setlength{\tabcolsep}{8pt}
\begin{tabular}{l cccc}
\hline
Training Strategy & OSR$\uparrow$ & NE$\downarrow$ & SR$\uparrow$ & SPL$\uparrow$ \\
\hline
SFT only & 87.48 & 2.34 & 79.40 & 69.97 \\
RL only & 74.52 & 4.30 & 64.30 & 59.82 \\
Sequential SFT-RL & 88.95 & 2.22 & 81.25 & 72.08 \\
Ours & \textbf{89.02} & \textbf{2.19} & \textbf{81.88} & \textbf{72.65} \\
\hline
\end{tabular}
\end{table}

%% file: sec/5_conclusion.tex
\section{Conclusion}

We present NavGRPO, a reinforcement learning framework for VLN that learns from diverse trajectories through Group Relative Policy Optimization. By comparing complete navigation rollouts, our method achieves stable policy updates without additional value networks. Experiments on several benchmarks show consistent improvements over imitation learning baselines, with substantial gains under perturbations, demonstrating that goal-directed RL training builds more robust navigation policies.

%% file: sec/X_suppl.tex
\clearpage
\setcounter{page}{1}
\maketitlesupplementary

\section{GRPO Variants Implementation}

We provide detailed formulations of the group-based policy optimization variants evaluated in our experiments. All variants share the same trajectory sampling strategy but differ in advantage estimation and policy update mechanisms.

\noindent\textbf{GRPO~\cite{guo2025deepseek}.}
Uses group-wise advantage normalization with both variance and length normalization:
\begin{equation}
\hat{A}_k = \frac{r_k - \text{mean}(\{r_1, ..., r_K\})}{\text{std}(\{r_1, ..., r_K\})}
\end{equation}
The policy objective aggregates with length normalization:
\begin{equation}
\mathcal{J}_{\text{GRPO}} = \mathbb{E}\left[ \frac{1}{K} \sum_{k=1}^{K} \frac{1}{|\tau_k|} \sum_{t=1}^{|\tau_k|} \rho_{k,t}\hat{A}_k \right]
\end{equation}
where $\rho_{k,t}(\theta) = \pi_\theta(a_{k,t}|s_{k,t})/\pi_{\text{old}}(a_{k,t}|s_{k,t})$ with clipping omitted for brevity. Variance normalization ensures consistent gradient scales across batches, while the $1/|\tau_k|$ term computes per-token average gradients. Token-level importance sampling applies independent ratio clipping at each time step.

\noindent\textbf{GSPO~\cite{zheng2025group}.}
Shifts importance sampling to sequence level. The sequence-level importance ratio is:
\begin{equation}
s_k(\theta) = \exp\left(\frac{1}{|\tau_k|}\sum_{t=1}^{|\tau_k|}\log\frac{\pi_\theta(a_{k,t}|s_{k,t})}{\pi_{\text{old}}(a_{k,t}|s_{k,t})}\right)
\end{equation}
The objective applies trajectory-level clipping:
\begin{equation}
\mathcal{J}_{\text{GSPO}} = \mathbb{E}\left[\frac{1}{K}\sum_{k=1}^{K} s_k\hat{A}_k\right]
\end{equation}
The sequence-level importance ratio $s_k$ aggregates token probabilities across the entire trajectory before clipping. This can be overly aggressive as it discards all tokens once triggered.

\noindent\textbf{GMPO~\cite{zhao2025geometric}.}
Replaces arithmetic mean with geometric mean for token-level rewards, computed in log space for numerical stability. The objective is:
\begin{equation}
\mathcal{J}_{\text{GMPO}} = \mathbb{E} \left[ \frac{1}{K}\sum_{k=1}^{K} \text{sgn}(\hat{A}_k) \cdot \left(\prod_{t=1}^{|\tau_k|}|\rho_{k,t}\hat{A}_k| \right)^{\frac{1}{|\tau_k|}}  \right]
\end{equation}
where clipping is omitted for brevity. The geometric mean inherently suppresses outliers, yielding more stable importance sampling ratios. GMPO employs token-level clipping with an exponential range $(e^{-\epsilon}, e^{\epsilon})$ that is wider than GRPO's linear range $(1-\epsilon, 1+\epsilon)$.

\section{Dr.GRPO for VLN}

\subsection{Background: Standard GRPO vs. Dr.GRPO}

Standard GRPO \cite{guo2025deepseek}, designed for language model alignment, applies two normalization terms to the advantage estimation:
\begin{equation}
\hat{A}^{\text{GRPO}}_k = \frac{1}{|\tau_k|} \cdot \frac{r_k - \text{mean}(\{r_1, ..., r_K\})}{\text{std}(\{r_1, ..., r_K\})}
\end{equation}
where $|\tau_k|$ is the trajectory length and $\text{std}(\{r_1, ..., r_K\})$ is the standard deviation of rewards across $K$ rollouts. While these normalizations are appropriate for open-ended text generation, they introduce optimization biases detrimental to navigation learning.

Dr.GRPO \cite{liu2025understanding} eliminates both normalizations through unbiased advantage estimation:
\begin{equation}
\hat{A}_k = r_k - \text{mean}(\{r_1, ..., r_K\})
\end{equation}
and aggregates gradients without normalization:
\begin{equation}
\mathcal{J}_{\text{Dr.GRPO}} = \mathbb{E}\left[\frac{1}{K} \sum_{k=1}^{K} \sum_{t=1}^{|\tau_k|} \rho_{k,t}\hat{A}_k \right]
\end{equation}
This recovers the theoretically grounded Monte Carlo policy gradient with an unbiased baseline.

\subsection{Why Standard GRPO Fails for VLN}

The two normalization terms in standard GRPO create specific problems for embodied navigation tasks:

\noindent\textbf{Length Normalization Bias.} The $1/|\tau_k|$ term divides the advantage by trajectory length to treat each token equally in language model training. However, in VLN, trajectory length naturally reflects scene complexity and navigation difficulty---longer paths may traverse larger buildings or require more precise localization. Empirically, R2R trajectories range from 5 to 15+ meters with an average of approximately 10 meters, exhibiting substantial variance. This creates asymmetric gradient updates:
\begin{itemize}
\item \textbf{Correct short paths} receive amplified gradients ($\propto 1/\text{short}$), encouraging shortcuts even when inappropriate.
\item \textbf{Incorrect long paths} receive diluted penalties ($\propto 1/\text{long}$), failing to discourage wandering behavior when the agent is lost.
\end{itemize}

\noindent\textbf{Variance Normalization Bias.} The $1/\text{std}(\{r_1, ..., r_K\})$ term amplifies gradients for instructions where all $K$ rollouts yield similar outcomes (low variance). This assigns disproportionate weight to:
\begin{itemize}
\item \textbf{'Easy' instructions} where all rollouts succeed (low learning value).
\item \textbf{'Hard' instructions} where all rollouts fail (no positive signal to exploit).
\end{itemize}
Both scenarios provide minimal actionable gradients, yet receive higher weight than instructions with mixed success/failure (high learning value). For VLN, where learning should prioritize instructions that distinguish successful from failed behaviors, this variance-based weighting is counterproductive.

By removing both normalizations, Dr.GRPO ensures that gradient magnitudes reflect true action importance rather than being artificially modulated by path length or scenario-specific variance, yielding more principled policy updates and improved sample efficiency.

\noindent Algorithm~\ref{alg:drgrpo} presents the complete training procedure.

\begin{algorithm}[t]
\caption{Dr.GRPO for VLN}
\label{alg:drgrpo}
\begin{algorithmic}[1]
\REQUIRE Pretrained policy $\pi_{\text{ref}}$, instruction set $\mathcal{D}$, Batch size $B$, group size $K$, hyperparameters $\alpha, \delta, \beta$
\STATE Initialize $\pi_\theta \leftarrow \pi_{\text{ref}}$, $\mathcal{B}_{\text{hard}} \leftarrow \emptyset$
\FOR{iteration $= 1, 2, ...$}
    \STATE Sample $\{W_i\}_{i=1}^B \sim \mathcal{D}$; $\pi_{\text{old}} \leftarrow \pi_\theta$
    \FOR{each $W_i$}
        \FOR{$k = 1$ to $K$}
            \STATE Sample $\tau_k \sim \pi_{\text{old}}(\cdot|W_i)$; compute $r_k$ (Eq.~8)
        \ENDFOR
        \STATE Compute $\hat{A}_k = r_k - \frac{1}{K}\sum_{j=1}^K r_j$ for all $k$
        \IF{$\sum_{k=1}^K \mathbb{I}(d_k < \epsilon) = 0$}
            \STATE $\mathcal{B}_{\text{hard}} \leftarrow \mathcal{B}_{\text{hard}} \cup \{W_i\}$
        \ENDIF
    \ENDFOR
    \STATE Update $\pi_\theta$ via gradient ascent (Eq.~5)
    \IF{$|\mathcal{B}_{\text{hard}}| \geq M$}
        \STATE SFT on $\mathcal{B}_{\text{hard}}$ (Eq.~11); $\mathcal{B}_{\text{hard}} \leftarrow \emptyset$
    \ENDIF
\ENDFOR
\RETURN $\pi_\theta$
\end{algorithmic}
\end{algorithm}

\section{Additional Experiments}

\subsection{Comparison with More RL Methods}

\input{tables/hamt}
\input{tables/alpha_ablation}

We compare with HAMT~\cite{chen2021hamt}, a representative value-based RL approach that applies policy gradient optimization with auxiliary value networks for vision-language navigation. We implement HAMT's training pipeline on ScaleVLN under identical training data and computational budget to ensure fair comparison.

Table~\ref{tab:hamt} demonstrates NavGRPO's clear advantage over HAMT across all settings. Under standard conditions, NavGRPO achieves +6.53\% SPL improvement. Under perturbations, NavGRPO exhibits substantially stronger robustness: degrading 1.6× less than HAMT under global perturbation and 1.5× less under early perturbation.

More revealing is the comparison across learning paradigms. While HAMT improves robustness over imitation learning, degrading roughly half as much as ScaleVLN under global perturbations, it suffers severe success rate collapse under extreme scenarios. HAMT's success rate drops 7.82\% under 2-step early perturbation compared to ScaleVLN's 3.71\% drop, revealing that traditional value-based RL struggles to maintain task completion capability when facing critical early errors.

The performance gap stems from two fundamental differences. First, value-based methods require training critic networks to approximate state values, which suffers from instability in VLN's high-dimensional action space. Second, traditional RL optimizes trajectories independently using absolute rewards, learning what is "correct." NavGRPO instead performs group-relative optimization—sampling diverse rollouts per instruction and learning through within-group comparison. This enables distinguishing relative quality among strategies under identical instructions, extracting supervision from both successful and failed trajectories that value-based methods cannot capture when optimizing trajectories in isolation.

\subsection{Extended Ablation Studies}

\noindent\textbf{Effect of Path Efficiency Weight $\alpha$.} 
Table~\ref{tab:alpha_ablation} demonstrates the impact of path efficiency weight $\alpha$ in $R_{\text{path}}$. Without path guidance at $\alpha=0$, the agent achieves highest success rate but suffers 1.17 SPL degradation due to inefficient detours, revealing the trade-off between task completion and path quality. Our default $\alpha=0.25$ strikes optimal balance, jointly optimizing both metrics. As $\alpha$ increases, performance degrades monotonically: $\alpha=1.0$ causes 1.8 SPL and 1.12 SR drops. This pattern reveals a limitation of excessive path efficiency enforcement. When deviations from oracle trajectories are heavily penalized, the agent develops brittle behavior that rigidly follows expected routes. In scenarios requiring adaptive replanning around obstacles or unexpected states, such rigidity leads to navigation failure rather than flexible recovery.

\input{tables/beta_ablation}
\input{tables/buffer_size}
\noindent\textbf{Robustness to Clipping and Entropy Regularization.}
We examine the sensitivity of our method to the entropy regularization coefficient $\beta$, as shown in Table~\ref{tab:beta_ablation}. Starting from a well-performing SFT policy, we find that a small amount of entropy regularization with $\beta = 0.01$ slightly improves performance compared to no regularization at $\beta = 0.0$. This can be attributed to the entropy term providing a mild regularization effect that prevents the policy from deviating too aggressively from the initial SFT policy during RL fine-tuning, thereby preserving the beneficial behaviors learned from supervised demonstrations. However, when $\beta$ is set too high at $0.1$, performance degrades significantly. Excessive entropy regularization overly restricts the policy's ability to explore and deviate from the SFT initialization, limiting the potential improvements that RL can achieve through reward-driven optimization. The relatively small performance gap between $\beta = 0.0$ and $\beta = 0.01$ suggests that our training framework strikes a good balance without requiring strong entropy constraints.Regarding the clipping threshold $\delta$, we do not conduct extensive ablation experiments as our training procedure accumulates gradients across all samples within a single rollout before performing a unified policy update. This ensures that the policy update remains on-policy, and the gradient accumulation naturally stabilizes the update magnitude without requiring aggressive clipping. We use the standard value $\delta = 0.2$ throughout our experiments.

\noindent\textbf{Hard Case Buffer Size.}
Table~\ref{tab:hard_case_buffer} shows the impact of hard case buffer size M. Performance remains stable across $M \in \{100, 200, 400\}$, with minimal variation in SR and SPL. This robustness stems from two factors: first, hard case updates are triggered only when all K sampled trajectories fail, making the actual update frequency relatively low regardless of buffer size; second, our GRPO training already maintains reasonable coverage of diverse navigation scenarios, reducing reliance on explicit hard case intervention. We include this mechanism primarily as a training stabilizer to prevent catastrophic forgetting on persistently challenging instructions, rather than as a core algorithmic component. We adopt M = 200 as a conservative choice that balances memory overhead and potential benefit without introducing additional hyperparameter sensitivity.

\section{Computational Cost Analysis}

\input{tables/cost}

\noindent Table~\ref{tab:computational_cost} presents a comprehensive computational cost comparison between ScaleVLN and NavGRPO on a single NVIDIA A800 GPU. NavGRPO with K=4 demonstrates superior memory efficiency, consuming only 14.788 GB of peak GPU memory compared to ScaleVLN's 20.274 GB, despite generating twice the number of rollouts per sample. This efficiency stems from our decoupled sampling-training architecture: the sampling phase computes visual feature embeddings and graph representations without maintaining computational graphs, caching only the processed features, while the training phase reconstructs gradients solely through the policy network using these cached representations, thereby eliminating redundant forward passes for visual feature computation and substantially reducing the memory burden of multi-trajectory optimization. The memory footprint scales linearly with rollout number K, doubling to 28.652 GB at K=8, representing a manageable and predictable scaling behavior. Training time per thousand steps increases moderately with K, yet our approach typically converges within 10k steps, maintaining reasonable total training cost while achieving improved performance. With K=8 achieving 72.65\% SPL compared to ScaleVLN's 69.97\%, NavGRPO demonstrates a favorable performance-efficiency trade-off that justifies the moderate computational overhead.

\section{Additional Visualizations}

\subsection{Qualitative Results}

\begin{figure}[t]
 \centering
\includegraphics[width=\linewidth]{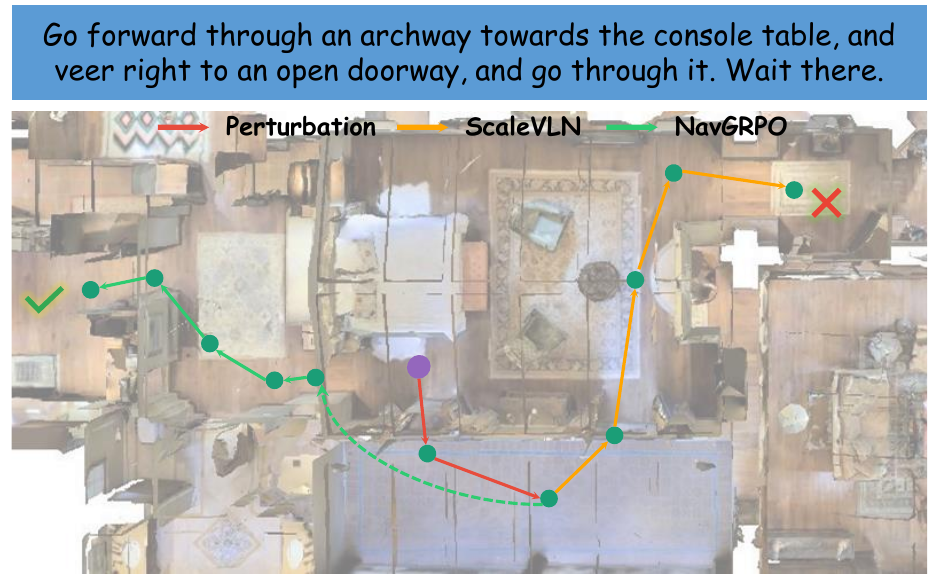}
\caption{Qualitative comparison of navigation trajectories. Purple point indicates the starting location. NavGRPO's trajectory (green) successfully completes the instruction by navigating through the archway and reaching the target doorway, while ScaleVLN's trajectory (yellow) fails to execute the complete route, demonstrating NavGRPO's superior robustness against environmental complexity.}
\label{fig:extend_result}
\end{figure}

\noindent Figure~\ref{fig:extend_result} presents a representative navigation example that highlights the robustness differences between methods. Starting from the purple initial point, the agent must navigate through the archway toward the console table and then veer right to reach the open doorway. The red trajectory represents an initial perturbation that could mislead the agent. NavGRPO demonstrates stronger robustness by recovering from potential distractions and adhering to the instructional semantics, while ScaleVLN shows greater sensitivity to such perturbations. This qualitative observation aligns with our quantitative results, confirming that multi-trajectory optimization with balanced rewards enhances policy stability in complex navigation scenarios.

\subsection{Failure Analysis}

\begin{figure}[t]
 \centering
\includegraphics[width=\linewidth]{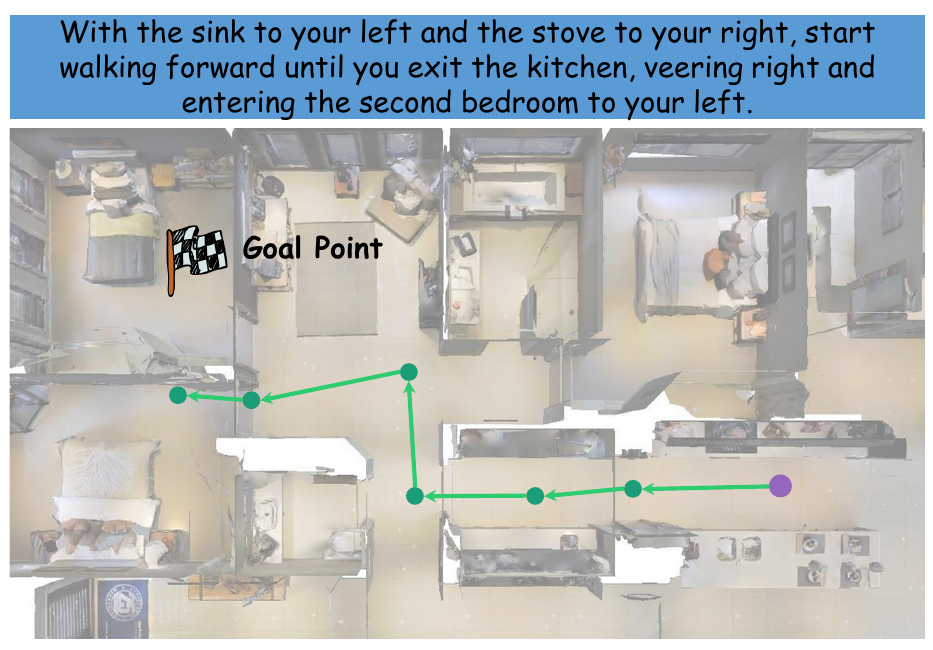}
\caption{Failure case analysis. NavGRPO incorrectly interprets the spatial reference "second bedroom to your left" in an instruction with multiple ambiguous directional cues, highlighting challenges in compositional spatial reasoning.}
\label{fig:failure_case}
\end{figure}

\noindent Figure~\ref{fig:failure_case} illustrates a challenging scenario where NavGRPO fails to reach the correct goal. The instruction requires the agent to navigate from the kitchen, veer right, and enter the second bedroom to the left. However, the spatial reference "second bedroom" is ambiguous given multiple visually similar bedroom entrances, and the conflicting directional cues "right" and "left" within the same instruction introduce additional complexity. The agent successfully exits the kitchen but misidentifies the target bedroom, revealing limitations in resolving compositional spatial relations across long-horizon instructions. This case underscores the challenge of grounding ambiguous linguistic spatial references in environments with repetitive structures, suggesting future directions in enhanced spatial reasoning and disambiguation mechanisms.

\section{Limitations}

Dependency on Base Model Quality. Our method fine-tunes pretrained vision-language models and inherits their limitations. When the base model exhibits systematic failures (e.g., misinterpreting certain spatial relations or object attributes), group-relative optimization may amplify rather than correct these biases, as all sampled trajectories suffer from the same foundational errors.


%% file: tables/hamt.tex
\begin{table}[t]
\centering
\caption{Comparison with value-based RL method (HAMT) under standard and perturbed settings on R2R Val Unseen. Both methods build upon ScaleVLN baseline.}
\label{tab:hamt}
\small
\setlength{\tabcolsep}{3.5pt}
\begin{tabular}{lccccc}
\toprule
Method & OSR↑ & NE↓ & SR↑ & SPL↑ & $\Delta$SPL \\
\midrule
\multicolumn{6}{l}{Standard (No Perturbation)} \\
\midrule
ScaleVLN + HAMT & 83.07 & 2.65 & 74.98 & 66.12 & -0.00 \\
ScaleVLN + NavGRPO & \textbf{89.18} & \textbf{2.19} & \textbf{81.88} & \textbf{72.65} & \textbf{-0.00} \\
\midrule
\multicolumn{6}{l}{Global Perturbation ($p=0.8$)} \\
\midrule
ScaleVLN + HAMT & 82.14 & 2.83 & 72.01 & 61.92 & -4.20 \\
ScaleVLN + NavGRPO & \textbf{88.49} & \textbf{2.27} & \textbf{79.95} & \textbf{69.98} & \textbf{-2.67} \\
\midrule
\multicolumn{6}{l}{Early Perturbation (Steps=2)} \\
\midrule
ScaleVLN + HAMT & 75.24 & 3.09 & 67.16 & 55.15 & -10.97 \\
ScaleVLN + NavGRPO & \textbf{83.85} & \textbf{2.70} & \textbf{76.50} & \textbf{65.10} & \textbf{-7.55} \\
\bottomrule
\end{tabular}
\end{table}

%% file: tables/alpha_ablation.tex
\begin{table}[t]
\centering
\caption{Ablation study on path efficiency weight $\alpha$ in $R_{\text{path}}$ on R2R Val Unseen split.}
\label{tab:alpha_ablation}
\small
\setlength{\tabcolsep}{14pt}
\begin{tabular}{l cccc}
\hline
$\alpha$ & OSR$\uparrow$ & NE$\downarrow$ & SR$\uparrow$ & SPL$\uparrow$ \\
\hline
0.0 & 88.64 & 2.26 & \textbf{81.92} & 71.48 \\
0.25 & \textbf{89.18} & \textbf{2.19} & 81.88 & \textbf{72.65} \\
0.5 & 88.71 & 2.23 & 81.34 & 71.92 \\
1.0 & 88.23 & 2.31 & 80.76 & 70.85 \\
\hline
\end{tabular}
\end{table}

%% file: tables/beta_ablation.tex
\begin{table}[t]
\centering
\caption{Ablation study on entropy regularization coefficient $\beta$ on R2R Val Unseen split.}
\label{tab:beta_ablation}
\small
\setlength{\tabcolsep}{14pt}
\begin{tabular}{l cccc}
\hline
$\beta$ & OSR$\uparrow$ & NE$\downarrow$ & SR$\uparrow$ & SPL$\uparrow$ \\
\hline
0.0  & 88.52 & 2.27 & 81.64 & 72.28 \\
0.01 & \textbf{89.18} & \textbf{2.19} & \textbf{81.88} & \textbf{72.65} \\
0.05 & 88.73 & 2.24 & 81.52 & 71.84 \\
0.1  & 88.21 & 2.32 & 80.92 & 70.56 \\
\hline
\end{tabular}
\end{table}

%% file: tables/buffer_size.tex
\begin{table}[t]
\centering
\caption{Impact of hard case buffer size M on R2R Val Unseen split. All variants substantially outperform Sequential SFT-RL baseline.}
\label{tab:hard_case_buffer}
\small
\setlength{\tabcolsep}{8pt}
\begin{tabular}{l cccc}
\hline
Training Strategy & OSR$\uparrow$ & NE$\downarrow$ & SR$\uparrow$ & SPL$\uparrow$ \\
\hline
Sequential SFT-RL & 88.95 & 2.22 & 81.25 & 72.08 \\
\hline
Ours (M=100)  & 88.89 & 2.23 & 81.64 & 72.41 \\
Ours (M=200)  & 89.18 & 2.19 & 81.88 & 72.65 \\
Ours (M=400)  & 89.02 & 2.20 & 81.77 & 72.53 \\
\hline
\end{tabular}
\end{table}

%% file: tables/cost.tex
\begin{table}[t]
\centering
\caption{Computational cost comparison between ScaleVLN and NavGRPO on R2R training.}
\label{tab:computational_cost}
\small
\setlength{\tabcolsep}{1pt}
\begin{tabular}{lccc}
\hline
& ScaleVLN & Ours(K=4) & Ours(K=8) \\
\hline
NE$\downarrow$ & 2.34 & 2.28 & 2.19 \\
SR$\uparrow$ & 79.40 & 80.37 & 81.88 \\
SPL$\uparrow$ & 69.97 & 71.98 & 72.65 \\
\hline
FLOPs(G) & 4.95 & 4.95 & 4.95 \\
Batchsize & 16 & 8 & 8 \\
Rollouts per Sample & 2 & 4 & 8 \\
Peak GPU Memory(GB)$\downarrow$ & 20.274 & 14.788 & 28.652 \\
Training Time(min/1k steps)$\downarrow$ & 46 & 51 & 86 \\
\hline
\end{tabular}
\end{table}